\documentclass{svproc}

\def\bng{\bngx}

\font\bngx=bang10

\def\*#1*#2{o\null{#2}{#1}}

\def\sh#1{\setbox0=\hbox{#1}%
     \kern-.02em\copy0\kern-\wd0
     \kern.04em\copy0\kern-\wd0
     \kern-.02em\raise.0433em\box0 }
\usepackage{tabularray}
\usepackage{float}
\usepackage{graphicx}
\usepackage{url}
\usepackage{xurl}
\usepackage[colorlinks=true, allcolors=blue]{hyperref}
\usepackage{subcaption}
\newcommand*\samethanks[1][\value{footnote}]{\footnotemark[#1]}
\newcommand{\hochkomma}{$^{,}$}

\begin{document}
\mainmatter              
\title{Rank Your Summaries: Enhancing Bengali Text Summarization via Ranking-based Approach}

\titlerunning{Enhancing Bengali Text Summarization via Ranking-based Approach}  
\author{G. M. Shahariar\thanks{These authors contributed equally to this work. Author names are in alphabetic order.}, Tonmoy Talukder\samethanks, Rafin Alam Khan Sotez, Md. Tanvir Rouf Shawon}
\authorrunning{Shahariar et al.} %
\institute{Ahsanullah University of Science and Technology, Dhaka, Bangladesh\\
\email{\{sshibli745,tonmoytalukder.cs,rafinkhan298,shawontanvir95\}@gmail.com}}

\maketitle              

\begin{abstract}
With the increasing need for text summarization techniques that are both efficient and accurate, it becomes crucial to explore avenues that enhance the quality and precision of pre-trained models specifically tailored for summarizing Bengali texts. When it comes to text summarization tasks, there are numerous pre-trained transformer models at one's disposal. Consequently, it becomes quite a challenge to discern the most informative and relevant summary for a given text among the various options generated by these pre-trained summarization models. This paper aims to identify the most accurate and informative summary for a given text by utilizing a simple but effective ranking-based approach that compares the output of four different pre-trained Bengali text summarization models. The process begins by carrying out preprocessing of the input text that involves eliminating unnecessary elements such as special characters and punctuation marks. Next, we utilize four pre-trained summarization models to generate summaries, followed by applying a text ranking algorithm to identify the most suitable summary. Ultimately, the summary with the highest ranking score is chosen as the final one. To evaluate the effectiveness of this approach, the generated summaries are compared against human-annotated summaries using standard NLG metrics such as \textit{BLEU, ROUGE, BERTScore, WIL, WER}, and \textit{METEOR}. Experimental results suggest that by leveraging the strengths of each pre-trained transformer model and combining them using a ranking-based approach, our methodology significantly improves the accuracy and effectiveness of the Bengali text summarization. 
\end{abstract}
\section{Introduction}
The process of text summarization involves condensing a given text while preserving its essential information and main ideas \cite{nenkova2012survey}. The increasing attention towards text summarization in the recent years is driven by the abundance of textual data available on the internet. With the exponential growth of digital content, it has become impractical for individuals to read every piece of information, which has led to the demand for automated methods to efficiently extract relevant information. Text summarization addresses this challenge by providing a concise representation of the original text, enabling users to effectively navigate through vast amounts of information. However, text summarization presents specific challenges for languages with limited resources, such as Bengali \cite{kumar2021study}.

Bengali, being a low-resource language, faces challenges due to the lack of annotated data and linguistic resources compared to more widely spoken languages like English \cite{uddin2007study}. Consequently, it becomes difficult to develop accurate and effective text summarization systems for Bengali as the availability of language-specific tools and models becomes limited \cite{ref12}.
However, the recent advances of the transformer based language models such as BERT (Bidirectional Encoder Representations from Transformers) \cite{ref15} and multilingual T5 (Text-to-Text Transfer Transformer) \cite{ref11} pre-trained on large amount of Bengali data have significantly improved Bengali text summarization task. 
By employing transfer learning \cite{pan2009survey}, pre-trained language models provide a foundation of linguistic knowledge and contextual understanding that helps boost the performance of Bengali text summarization systems. 

Even with the availability of pre-trained models, selecting the most suitable summary remains a challenge. Numerous pre-trained transformer models and deep learning models claim to offer optimal outcomes for Bengali text summarization \cite{liu-lapata-2019-text}. However, due to the subjective nature of summarization, it is difficult to determine an objective ``best" summary. To address this challenge and enhance the accuracy and quality of the summarization task, a rank-based approach can be utilized. Unlike existing summarization models that generate a single output for a given text, our model can identify the most appropriate summary by comparing multiple summaries generated by different pre-trained models for the same Bengali input text.

In this study, we present a simple yet effective approach that utilizes multiple models to generate candidate summaries and then rank them based on their quality. For experimentation, we used four pre-trained Bengali text summarization models to generate four distinct candidate summaries for a given input text. Then, we employed a graph-based technique to rank the candidate summaries and ultimately selected the one with the highest ranking score which we refer to as the most suitable summary. This method allows for a comprehensive evaluation of the generated summaries, enabling us to identify the most informative and coherent one. The four pre-trained models we considered in this study consisted of both extractive and abstractive Bengali text summarization models. To assess and compare the performance of our proposed approach, we computed various scores between the candidate and reference summaries. These scores include Word Error Rate (WER), Word Information Lost (WIL), Metric for Evaluation of Translation with Explicit ORdering (METEOR), BERTScore, Recall-Oriented Understudy for Gisting Evaluation (ROUGE-1, ROUGE-2, ROUGE-L), and BiLingual Evaluation Understudy (BLEU-3, BLEU-4). In summary, we have made the following contributions in this paper:
\begin{enumerate}
    \item We introduce a simple yet effective approach of ranking the candidate summaries generated by multiple pre-trained transformer models, selecting the most suitable one based on its high ranking score. This method allows us to identify summaries that are both informative and coherent.
    \item To evaluate the effectiveness of our approach, we employ multiple metrics such as WER, WIL, METEOR, BERTScore, ROUGE-1, ROUGE-2, ROUGE-L, BLEU-3, and BLEU-4. These metrics provide valuable insights into the quality and fidelity of the candidate summaries when compared to the reference summaries.
    \item We have made our implementation of the proposed approach publicly available\footnote{\url{https://github.com/TonmoyTalukder/Rank-Your-Summaries-Enhancing-Bengali-Text-Summarization-via-Ranking-based-Approach}} with the aim of fostering collaboration and encouraging researchers to further enhance and contribute to the Bengali natural language generation field.
\end{enumerate}

\section{Bengali Summary Ranker}
In this section, we provide a comprehensive discussion of our detailed approach to Bengali text summarization using a ranking-based method.
\subsection{Proposed Approach}
\begin{figure*}[h!]
    \centering
    \includegraphics[scale=0.65]{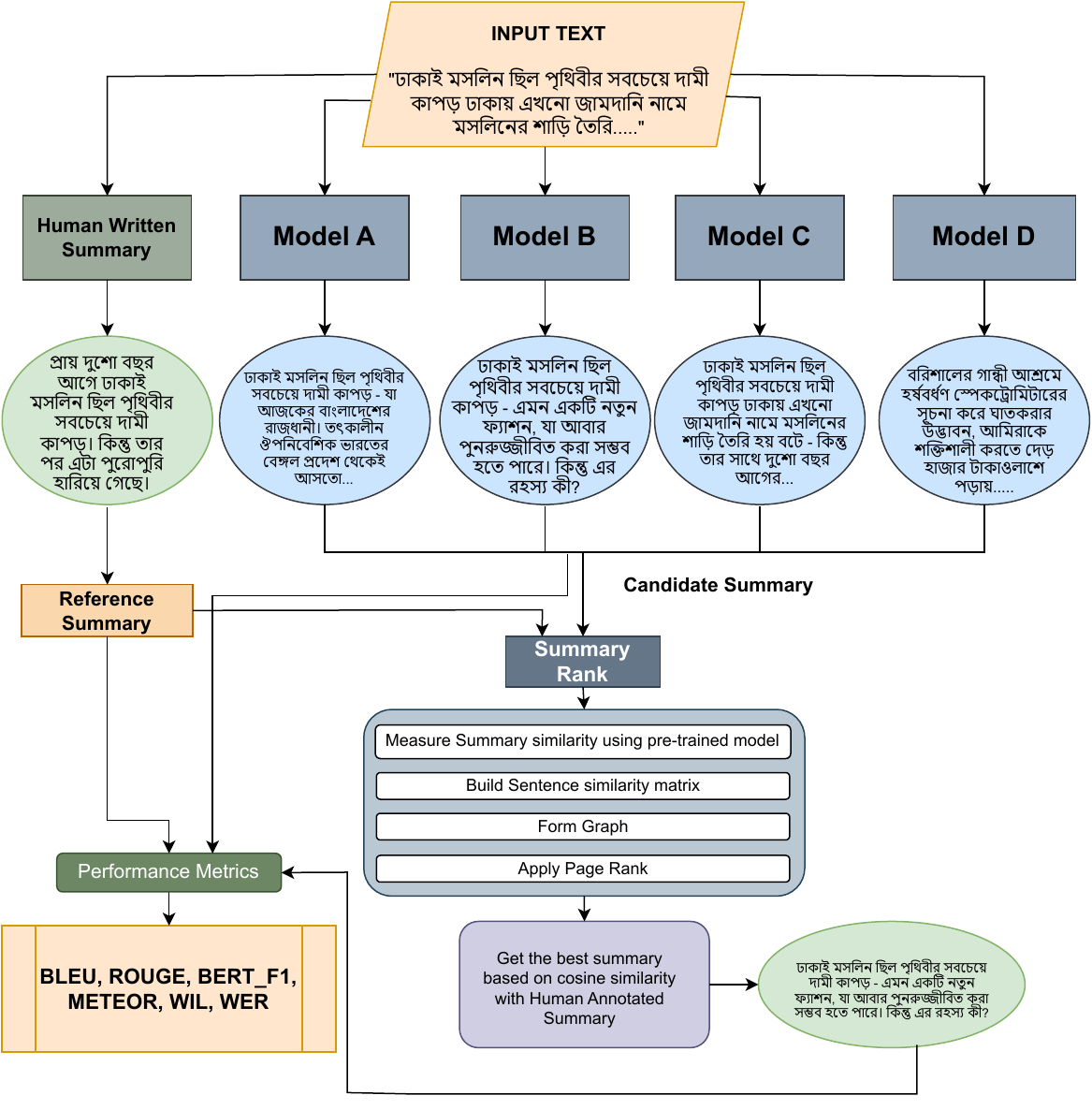}
    \caption{Schematic diagram of the proposed methodology. Here, model A, B, C, D are \textit{mt5 XLSum, mT5 CrossSum, scibert uncased}, and \textit{mT5 by shahidul} respectively.}
    \label{fig:Rankerimage}
\end{figure*}

To summarize a single text, we employ four pre-trained Bengali text summarization models. Subsequently, we utilize a graph-based approach to rank the candidate summaries and select the highest-ranked summary as the final output. The method we propose takes Bengali texts and their corresponding human-written summaries as input and provides the best summary from the four candidate summaries. The proposed methodology is depicted in Fig. \ref{fig:Rankerimage} and explained in details below.

\noindent \textbf{Step 1) Input text:} Each Bengali text from the dataset is sequentially processed through the proposed methodology for summarization.\\
\noindent \textbf{Step 2) Get the reference summary:} We acquire the corresponding summary (human written) for each text in the dataset. We refer the summary as `reference' summary.\\
\textbf{Step 3) Summarization:} We generate four candidate summaries for each text in the dataset by utilizing four different pre-trained text summarization models. The details of the text summarization models can be found in sub-section \ref{models}. \\
\textbf{Step 4) Ranking:} To determine the `best' summary among the four candidate summaries, we apply \verb|TextRank| \cite{b15} which is a graph-based ranking method. It utilizes a similarity matrix to gauge the level of similarity among summaries and ranks them by constructing a graph using \verb|PageRank| algorithm \cite{b16}. To represent candidate summaries in a network, a similarity matrix is calculated with the summaries serving as vertices and the connections between them serving as edges indicate similarity scores. The proposed approach for choosing the best summary involves evaluating the similarity of all candidate summaries with the reference summary (human written). We select the top-ranked summary as the best one. We incorporate the approach proposed in \cite{ref2} for ranking candidate summaries which is described as follows:\\
\textbf{\textit{(a) Summary similarity assessment}}: To evaluate the similarity between candidate summaries, we utilized the generator checkpoint of BanglaBERT \cite{ref16}, which is a pre-trained BERT based Bengali language model. We used this to embed the essential meaning of words in closely-bound vectors, effectively transforming each sentence into a vector where each value inside the vector holds a specific purpose. To obtain a contextual sentence embedding using BanglaBERT, we applied mean pooling to the last\_hidden\_state tensor, which results in 400 numeric values for the reference summary and each of the candidate summaries. To determine the similarity score between a reference and a candidate summary, we computed cosine similarity \cite{b18}.\\
\textbf{\textit{(b)	Summary similarity matrix calculation}}:  We created a $5 \times 5$ similarity matrix, which we refer to as the Summary Similarity Matrix (\verb|SSM|). The indices within the matrix are used to denote different summaries, with index 0 representing the human-written reference summary, and indices 1 to 4 representing the candidate summaries generated by the models. Initially, we set the matrix values to all zeros. Later, we assigned similarity scores to only the $0^{th}$ row and column since we were measuring the similarity of all the candidate summaries with the human-written summary, not among the candidate summaries themselves.\\
\textbf{\textit{(c)	Graph building and ranking}}: We constructed a graph using the \verb|SSM| matrix, where the vertices of the graph were the summaries, and the connections between them indicated their cosine similarity scores. To construct the graph and implement the \verb|PageRank| algorithm, we used NetworkX \footnote{\url{https://github.com/networkx/networkx}} python library. Ultimately, we ranked the candidate summaries based on their \verb|PageRank| value and selected the summary with the highest value as our final output. For the example text used in figure \ref{fig:Rankerimage}, {\bng ``DhakaI mosiln ichlo prRithbiir sobecey damii kaporh Dhakay Ekhena jamdain naem mosilenr shairh oitir...''} (truncated due to lengthy text) \textit{[Dhakai Muslin was the most expensive cloth in the world, still muslin sarees called Jamdani are made in Dhaka.....]}, the human written summary from the dataset is - {\bng ``pRay duesha bochor Aaeg DhakaI mosiln ichlo prRithbiir sobecey damii kaporh. ikn/tu tar por ETa puerapuir Hairey egech.''} \textit{[About two hundred years ago, Dhakai Muslin was the most expensive fabric in the world. But after that it was completely lost.]} and the best summary selected by our approach is -
{\bng ``DhakaI mosiln ichlo prRithbiir sobecey damii kaporh - Emon EkiT notun fYashn, Ja Aabar punruj/jiiibt kora sm/vb Het paer. ikn/tu Er rHsY kii?''} \textit{[Dhakai Muslin was the most expensive fabric in the world - a new fashion that may be revived. But what is the secret?]}
\\
\textbf{Step 5) Evaluation:} We measure the effectiveness of the proposed approach by comparing reference and best ranked summary as well as best ranked summary and the original text from the dataset. A wide range of evaluation metrics including BLEU–3, BLEU–4, ROUGE–1, ROUGE–2, ROUGE–L, BERTScore (F1), WIL, WER, and METEOR were used to conduct the performance evaluation.

\subsection{Models}\label{models}
Pre-trained text summarization models utilized in this study can be categorized into two broad categories: \textit{extractive} and \textit{abstractive}.\\
\noindent\textbf{(a) Abstractive Text Summarizer:}
Abstractive summarization generates summary that goes beyond the original text by rephrasing the original content into a new and concise summary. \verb|Multilingual T5 (mT5)| \cite{ref11}, a pre-trained language model utilizing the transformer architecture, is capable of performing a wide range of NLG tasks across multiple languages, including text-to-text generation, machine translation, summarization etc. We utilized three mT5 variants for abstractive summarization in this study. Hasan et al. \cite{ref12} fine-tuned mT5 on \textit{XLSum} dataset for summarization task on 45 languages. We refer to this model as \textit{`mT5 XLSum'} in this study. Bhattacharjee et al. \cite{ref13} fine-tuned mT5 on all cross-lingual pairs of the \textit{CrossSum} dataset to generate an abstract summary of any text in a specified target language. We refer to this model as \textit{`mT5 CrossSum'}. Another fine-tuned version of the mT5 is available at huggingface.co\footnote{\url{https://huggingface.co/shahidul034/Bangla\_text\_summarization\_model}} that uses \textit{MT5ForConditionalGeneration} model architecture. We refer to this model as \textit{`mT5 by shahidul'} for the rest of the study.\\
\noindent\textbf{(b) Extractive Text Summarizer:}
Extractive summarization involves finding and extracting the most relevant phrases and sentences from the original text to create a concise and coherent summary. We utilized \verb|SciBERT| \cite{ref14}, a pre-trained extractive summarizer built on top of the BERT architecture. We refer to this summarizer as \textit{`scibert uncased'} in this study. \verb|SciBERT| is specifically trained on the scientific text and is designed to perform well on multilingual NLG tasks related to scientific domains, such as scientific document retrieval, named entity recognition, extractive text summarization, and question answering.

\section{Evaluation}
\subsection{Datasets}\label{datasets}
We used two datasets for this study: \textit{XL-Sum Multilingual dataset} \cite{ref12} available on Huggingface datasets\footnote{\url{ https://huggingface.co/datasets/csebuetnlp/xlsum}} and \textit{Bangla Text Summarization dataset}\footnote{\url{https://www.kaggle.com/datasets/hasanmoni/bengali-text-summarization}}. The XL-Sum dataset is extensive and diverse, containing 1.35 million pairs of well-documented articles and summaries from the BBC, which were obtained utilizing some best-designed algorithms. The dataset includes 45 languages with varying levels of resources required, some of which do not have any existing accessible data sets. However, as Bengali is our primary ambition, we used the `Bengali' subset of the XL-Sum dataset. The XL-Sum dataset has 10126 Bengali summaries which are divided into the following splits: train 8102, validation 1012, and test 1012. The `Bangla Text Summarization dataset' has 80.3K data with the attributes named category, summary, and text. Although the dataset has 80.3K data, we have used 5K data chosen randomly.

\subsection{Hyper-parameter Settings}
We set the maximum output token length to 400 and minimum to 64 for both the abstractive and extractive summarizers. The maximum length of the input tokens was set to 512. During generation, to avoid repeating of same word multiple times, we have set \textit{no\_repeat\_ngram\_size} to 2. To introduce diversity, we used beam search with a \textit{beam size} of 4.  

\subsection{Evaluation Metrics}
\label{sec_eval}
\textbf{(a) BLEU Score:} Bilingual Evaluation Understudy (BLEU) \cite{b19} is a commonly used metric for evaluating the quality of machine-generated summaries. It measures the degree of overlap between n-grams in generated and reference summaries, where higher scores indicate better quality. In particular, BLEU-3 and BLEU-4 are variants of the BLEU metric that focus on tri-grams and quad-grams respectively, to provide a more detailed evaluation of the summarization quality.\\ 
\textbf{(b) ROUGE:} Recall-Oriented Understudy for Gisting Evaluation (ROUGE) \cite{b20} is a family of metrics used to evaluate the text summarization quality. ROUGE measures the overlap of n-grams between the generated and reference summary. ROUGE-1, ROUGE-2 and ROUGE-L refer to the uni-grams, bi-grams, and longest common sub-sequences based evaluation scores between two texts respectively.\newline
\textbf{(c) BERTScore:} BERTScore \cite{b21} is an automatic evaluation metric for text generation based on the BERT architecture. It calculates the F1 measurement of the BERT embeddings between the generated and reference summary, considering their precision and recall. BERTScore takes into account the semantic meaning of words and the context in which they are used, making it effective for evaluating text generation tasks. \newline
\textbf{(d) METEOR:} Metric for Evaluation of Translation with Explicit Ordering (METEOR) \cite{b23} performs evaluation based on a generalized concept of uni-gram matching between the generated and reference summary. METEOR can handle variations in word order and word choice during evaluation.\newline
\textbf{(e) WER:} Word Error Rate (WER) \cite{b24} is commonly used to evaluate the quality of machine-generated texts. WER is calculated by comparing the machine-generated summaries to the human-generated summary and counting the number of errors, such as insertions, deletions, and substitutions. It calculates how many ``errors'' are in the generated summary text, compared to a reference summary text.\newline
\textbf{(f) WIL:} Word Information Lost (WIL) \cite{b24} measures the amount of information lost in the generated summaries compared to the reference summary by computing the difference in entropy between the two texts. WIL Score is calculated by dividing the amount of information or content lost in the generated summary by the total amount of information contained in the reference summary.

\subsection{Experimental Results}
This section presents the performance evaluation of the proposed approach consisting of four pre-trained text summarization models. We assessed the performance using the standard NLG evaluation metrics discussed above. We have performed experimentations on the two datasets as discussed in \ref{datasets}. We performed two types of experimentations on both the datasets: (a) performance comparison between the input text and all the summaries (reference, generated and best ranked) which we report in Table \ref{tab:score_givenText}, and (b) performance comparison between the reference summary and all the generated summaries along with best ranked summary which is reported in Table \ref{tab:score_givenSummary} and \ref{tab:rogue_bleu}. All the reported scores are averaged over a whole dataset. 


\begin{table}[t]
\centering
\caption{Performance comparison between the input text and all the summaries on two different datasets. `Best Summary' is the summary selected by our approach and `Given Summary' is the human-written reference summary. Here, model A, B, C, and D are \textit{mt5 XLSum, mt5 CrossSum, scibert uncased} and \textit{mT5 by shahidul} respectively.}
\label{tab:score_givenText}
\resizebox{\textwidth}{0.68\height}
{\begin{tblr}{
  cells = {c},
  cell{1}{1} = {r=2}{},
  cell{1}{2} = {c=4}{},
  cell{1}{6} = {c=4}{},
  vline{1-2,6,10} = {1}{},
  vline{1-10} = {2}{},
  vline{1-2,6,10} = {3-9}{},
  hline{1,3-4,6-10} = {-}{},
  hline{2} = {2-9}{},
}
\textbf{Summary}       & \textbf{XLSum Dataset}      &                &                    &                & {\textbf{Bangla Text Summarization }\\\textbf{Dataset}} &                &                    &                \\
              & \textbf{WIL} & \textbf{METEOR}         & \textbf{WER} & {\textbf{BERTScore}\\\textbf{(F1)}}        & \textbf{WIL}                & \textbf{METEOR}         & \textbf{WER}  & {\textbf{BERTScore}\\\textbf{(F1)}}        \\
\textbf{Given Summary} & 0.0099              & 0.196          & 0.0098              & 0.673          & 0.0098                             & 0.278          & 0.0097              & 0.651          \\
& & & & & & & & \\
{\textbf{Best }\\\textbf{Summary}}   & \textbf{0.0095}     & \textbf{0.347} & \textbf{0.0094}     & \textbf{0.723} & \textbf{0.0092}                    & \textbf{0.361} & \textbf{0.0090}     & \textbf{0.725} \\
\textbf{Model A}       & 0.0098              & 0.320          & 0.0097              & 0.716          & 0.0095                             & 0.332          & 0.0092              & 0.715          \\
\textbf{Model B}       & 0.0098              & 0.296          & 0.0097              & 0.709          & 0.0095                             & 0.326          & 0.0093              & 0.714          \\
\textbf{Model C}       & \textbf{0.0081}     & 0.579          & \textbf{0.0081}     & \textbf{0.625} & \textbf{0.0082}                    & \textbf{0.489} & \textbf{0.0079}     & \textbf{0.765} \\
\textbf{Model D}       & 0.0100              & 0.025          & 0.0099              & 0.625          & 0.0099                             & 0.032          & 0.0098              & 0.624          
\end{tblr}}
\end{table}

\begin{table}[h]
\centering
\caption{Performance comparison between the reference and all other summaries (candidate and best ranked) on two different datasets. `Best Summary' is the summary selected by our approach. Here, model A, B, C, and D are \textit{mt5 XLSum, mt5 crosssum, scibert uncased} and \textit{mT5 by shahidul} respectively.}
\label{tab:score_givenSummary}
\resizebox{\textwidth}{!}
{\begin{tblr}{
  cells = {c},
  cell{1}{1} = {r=2}{},
  cell{1}{2} = {c=4}{},
  cell{1}{6} = {c=4}{},
  vline{1-2,6,10} = {1}{},
  vline{1-10} = {2}{},
  vline{1-2,6,10} = {3-8}{},
  hline{1,3,5-9} = {-}{},
  hline{2} = {2-9}{},
}
\textbf{Summary}       & \textbf{XLSum Dataset}      &                &                    &                & {\textbf{Bangla Text Summarization }\\\textbf{Dataset}} &                &                    &                \\
              & \textbf{WIL} & \textbf{METEOR}         & \textbf{WER} & {\textbf{BERTScore }\\\textbf{(F1)}}        & \textbf{WIL}                & \textbf{METEOR}         & \textbf{WER} & {\textbf{BERTScore }\\\textbf{(F1)}}        \\
& & & & & & & & \\
{\textbf{Best }\\\textbf{Summary}} & \textbf{0.0095}     & \textbf{0.189} & \textbf{0.017}     & \textbf{0.749} & \textbf{0.0094}                    & \textbf{0.192} & \textbf{0.040}     & \textbf{0.708} \\
\textbf{Model A}     & 0.0095              & 0.182          & 0.012              & \textbf{0.750} & 0.0095                             & 0.164          & \textbf{0.031}     & 0.701          \\
\textbf{Model B}     & 0.0097              & 0.143          & 0.012              & 0.735          & 0.0095                             & 0.163          & \textbf{0.031}     & 0.702          \\
\textbf{Model C}     & 0.0099              & 0.108          & 0.051            & 0.679          & 0.0096                             & 0.185          & 0.078              & 0.681          \\
\textbf{Model D}     & 0.0100              & 0.007          & 0.019              & 0.619          & 0.0099                             & 0.033          & 0.052              & 0.635          
\end{tblr}}
\end{table}


\section{Result Analysis}
In this section, we provide an analysis of the experimental results. Upon examining the findings, we made several intriguing observations.

\begin{table}[h]
\centering
\caption{BLEU and ROUGE scores comparison between the reference and all other summaries (candidate and best ranked) on two different datasets. Here, model A, B, C, D are \textit{mT5 XLSum, mT5 CrossSum, scibert uncased}, and \textit{mT5 by shahidul} respectively.}
\label{tab:rogue_bleu}
\resizebox{\textwidth}{0.68\height}
{\begin{tblr}{
  cells = {c},
  cell{2}{1} = {r=15}{},
  cell{2}{2} = {r=3}{},
  cell{2}{3} = {r=3}{},
  cell{2}{4} = {r=3}{},
  cell{5}{2} = {r=3}{},
  cell{5}{3} = {r=3}{},
  cell{5}{4} = {r=3}{},
  cell{8}{2} = {r=3}{},
  cell{8}{3} = {r=3}{},
  cell{8}{4} = {r=3}{},
  cell{11}{2} = {r=3}{},
  cell{11}{3} = {r=3}{},
  cell{11}{4} = {r=3}{},
  cell{14}{2} = {r=3}{},
  cell{14}{3} = {r=3}{},
  cell{14}{4} = {r=3}{},
  cell{17}{1} = {r=15}{},
  cell{17}{2} = {r=3}{},
  cell{17}{3} = {r=3}{},
  cell{17}{4} = {r=3}{},
  cell{20}{2} = {r=3}{},
  cell{20}{3} = {r=3}{},
  cell{20}{4} = {r=3}{},
  cell{23}{2} = {r=3}{},
  cell{23}{3} = {r=3}{},
  cell{23}{4} = {r=3}{},
  cell{26}{2} = {r=3}{},
  cell{26}{3} = {r=3}{},
  cell{26}{4} = {r=3}{},
  cell{29}{2} = {r=3}{},
  cell{29}{3} = {r=3}{},
  cell{29}{4} = {r=3}{},
  vline{-} = {1}{},
  vline{1-3,5-6,9} = {2-31}{},
  vline{6,9} = {3-4,6-7,9-10,12-13,15-16,18-19,21-22,24-25,27-28,30-31}{},
  vline{3,5-6,9} = {5,8,11,14,20,23,26,29}{},
  hline{1-2, 17, 32} = {-}{},
  hline{3-4,6-7,9-10,12-13,15-16,18-19,21-22,24-25,27-28,30-31} = {5-8}{},
  hline{5,8,11,14,17,20,23,26,29} = {2-8}{},
}
\textbf{Dataset}                                         & {\textbf{Summary }\\\textbf{Model}} & {\textbf{BLEU }\\\textbf{3}} & {\textbf{BLEU }\\\textbf{4}} & {\textbf{ROUGE }\\\textbf{Version}} & \textbf{Recall}  & \textbf{Precision} & {\textbf{F1}\\\textbf{Score}} \\
\textbf{XL-Sum}                                          & {\textbf{Best }\\\textbf{Summary}}  & \textbf{0.0783}              & \textbf{0.0496}              & $r$-$1$                                 & \textbf{0.313}   & \textbf{0.222}     & \textbf{0.249}                \\ &                                     &                              &                              & r$-$2                                 & \textbf{0.132}   & \textbf{0.096}     & \textbf{0.107}                \\ &                                     &                              &                              & r$-$l                                 & \textbf{0.260}   & \textbf{0.186}     & \textbf{0.208}                \\ & \textbf{Model A}                    & 0.0765                       & 0.0463                       & r$-$1                                 & 0.288            & \textbf{0.227}     & 0.245                         \\ &                                     &                              &                              & r$-$2                                 & 0.125            & \textbf{0.096}     & 0.105                         \\ &                                     &                              &                              & r$-$l                                 & 0.245            & \textbf{0.191}     & \textbf{0.208}                \\ & \textbf{Model B}                    & 0.0502                       & 0.029                        & r$-$1                                 & 0.235            & 0.187              & 0.201                         \\ &                                     &                              &                              & r$-$2                                 & 0.088            & 0.068              & 0.074                         \\ &                                     &                              &                              & r$-$l                                 & 0.197            & 0.155              & 0.168                         \\ & \textbf{Model C}                    & 0.0125                       & 0.0064                       & r$-$1                                 & 0.277            & 0.075              & 0.112                         \\ &                                     &                              &                              & r$-$2                                 & 0.072            & 0.018              & 0.027                         \\ &                                     &                              &                              & r$-$l                                 & 0.202            & 0.055              & 0.082                         \\ & \textbf{Model D}                    & 1.13E-05                     & 2.91E-82                     & r$-$1                                 & 0.017            & 0.010              & 0.012                         \\ &                                     &                              &                              & r$-$2                                 & 0.001            & 0.000              & 0.000                         \\ &                                     &                              &                              & r$-$l                                 & 0.016            & 0.010              & 0.012                         \\
{\textbf{Bangla}\\\textbf{Text}\\\textbf{Summarization}} & \textbf{Best Summary}  & \textbf{0.0300} & \textbf{0.0130} & r$-$1           & \textbf{0.433} & \textbf{0.118} & \textbf{0.184} \\&                        &                 &                 & r$-$2           & \textbf{0.176} & \textbf{0.044} & \textbf{0.069} \\&                        &                 &                 & r$-$l           & \textbf{0.392} & \textbf{0.107} & \textbf{0.167} \\& \textbf{Model A}       & 0.0253          & 0.0108          & r$-$1           & 0.369          & 0.107          & 0.165          \\&                        &                 &                 & r$-$2           & 0.144          & 0.038          & 0.060          \\&                        &                 &                 & r$-$l           & 0.337          & 0.098          & 0.151          \\& \textbf{Model B}       & 0.0248          & 0.0102          & r$-$1           & 0.367          & 0.108          & 0.166          \\&                        &                 &                 & r$-$2           & 0.141          & 0.038          & 0.059          \\&                        &                 &                 & r$-$l           & 0.332          & 0.098          & 0.151          \\& \textbf{Model C}       & 0.0200          & 0.0089          & r$-$1           & \textbf{0.454} & 0.080          & 0.132          \\&                        &                 &                 & r$-$2           & \textbf{0.187} & 0.029          & 0.049          \\&                        &                 &                 & r$-$l           & \textbf{0.415} & 0.073          & 0.121          \\& \textbf{Model D}       & 0.0009          & 0.0001          & r$-$1           & 0.099          & 0.020          & 0.034          \\&                        &                 &                 & r$-$2           & 0.016          & 0.003          & 0.005          \\&                        &                 &                 & r$-$l           & 0.093          & 0.019          & 0.032          
\end{tblr}}
\end{table}

\subsection{Quantitative Analysis}

\noindent\textbf{(a) Comparison with the given text:}
In Table \ref{tab:score_givenText}, the scores illustrate the performance comparison between the input text and all the summaries (reference, generated and best ranked) on two different datasets. When analyzing the \textit{XLSum} dataset, we observe that the performance of \textit{mT5 XLSum, mT5 CrossSum}, and \textit{mT5 by shahidul} models is relatively lower, despite their ability to provide abstractive summaries. The mismatch between the generated summaries and the original text, which occurs due to abstract generation, leads to less noteworthy performance. However, the scores indicate that \textit{scibert uncased} performs better than even the top-ranked summary. This could be attributed to the fact that \textit{scibert uncased} is an extractive summarizer, and our scoring is based on comparing the summaries with the input text. Unlike abstractive summarization, an extractive summary extracts words or sentences directly from the input text. As a result, summaries generated by the \textit{scibert uncased} model score higher due to increased word matching with the reference text compared to the other models. Similar observations can be made with the \textit{Bangla Text Summarization} dataset. In this case as well, the best summary produced by our proposed approach outperforms the individual abstractive summarizers but falls short of surpassing the extractive summarizer (\textit{scibert uncased}).

\begin{figure*}[h]
    \centering
    \includegraphics[scale=0.45]{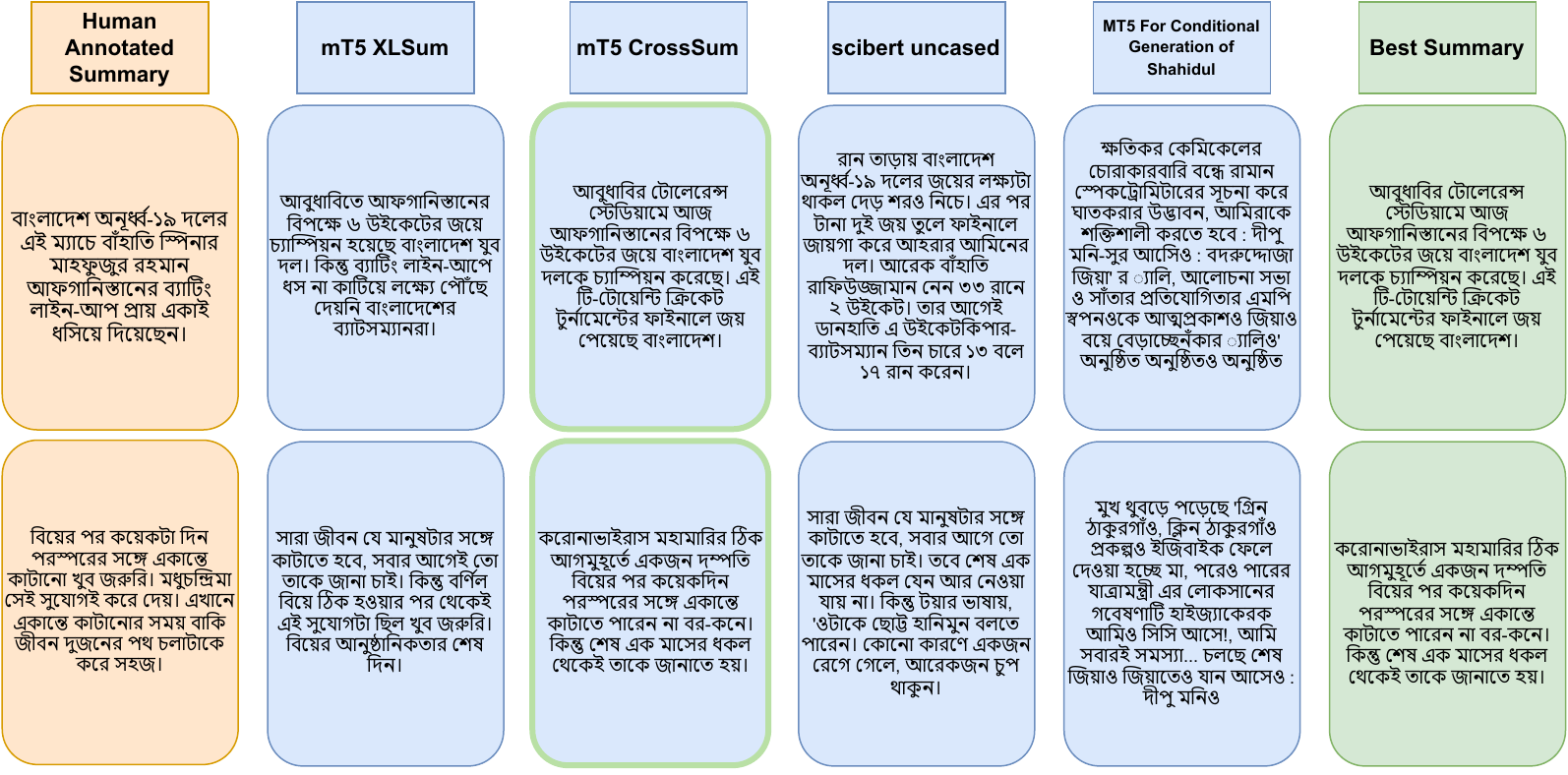}
    \caption{Example of a few candidate summaries generated by all the models along with the reference and best-ranked summary on two randomly picked newspaper texts.}
    \label{fig:BestSummaryExample}
\end{figure*}

\noindent\textbf{(b) Comparison with the reference summary:}
We conducted a thorough comparison of all the generated summaries with the human-annotated summaries. The results of this comparison can be seen in Table \ref{tab:score_givenSummary}, where our proposed approach achieved the highest BERTScore (F1 measure) of 0.749 and the lowest WIL score of 0.0095 for the \textit{XL-Sum} dataset. Similarly, for the other dataset, our approach consistently delivered the best results across various metrics. In contrast, the individual summarizers performed poorly in terms of performance scores. However, by utilizing our approach, we were able to achieve a notable improvement in the overall scores, as evidenced by the results. It is worth mentioning that although the BERTScore (F1-measure) is the second best for our best ranked summary, the \textit{mT5 XLSum} summarizer performs the best in this regard. This can be attributed to the fact that we utilized the \textit{XL-Sum} dataset on which the \textit{mT5 XLSum} summarizer was trained on. To validate the generalization performance of the summarizers, we tested two randomly chosen texts\footnote{\url{ https://www.prothomalo.com/sports/cricket/l4ihz4td0k}}\hochkomma\footnote{\url{https://www.prothomalo.com/lifestyle/travel/0pr5imen9q}} from a daily newspaper that were not present in either of the datasets. We compared the generated summaries from the summarizers with the human-annotated summaries of these texts (written by the second author) which is illustrated in Figure \ref{fig:BestSummaryExample}. Interestingly, our proposed approach ranked the summary generated by the \textit{mT5 CrossSum} model as the best in this particular comparison. 
\begin{figure*}[h]
\begin{center}
    \centering
    \includegraphics[scale=0.55]{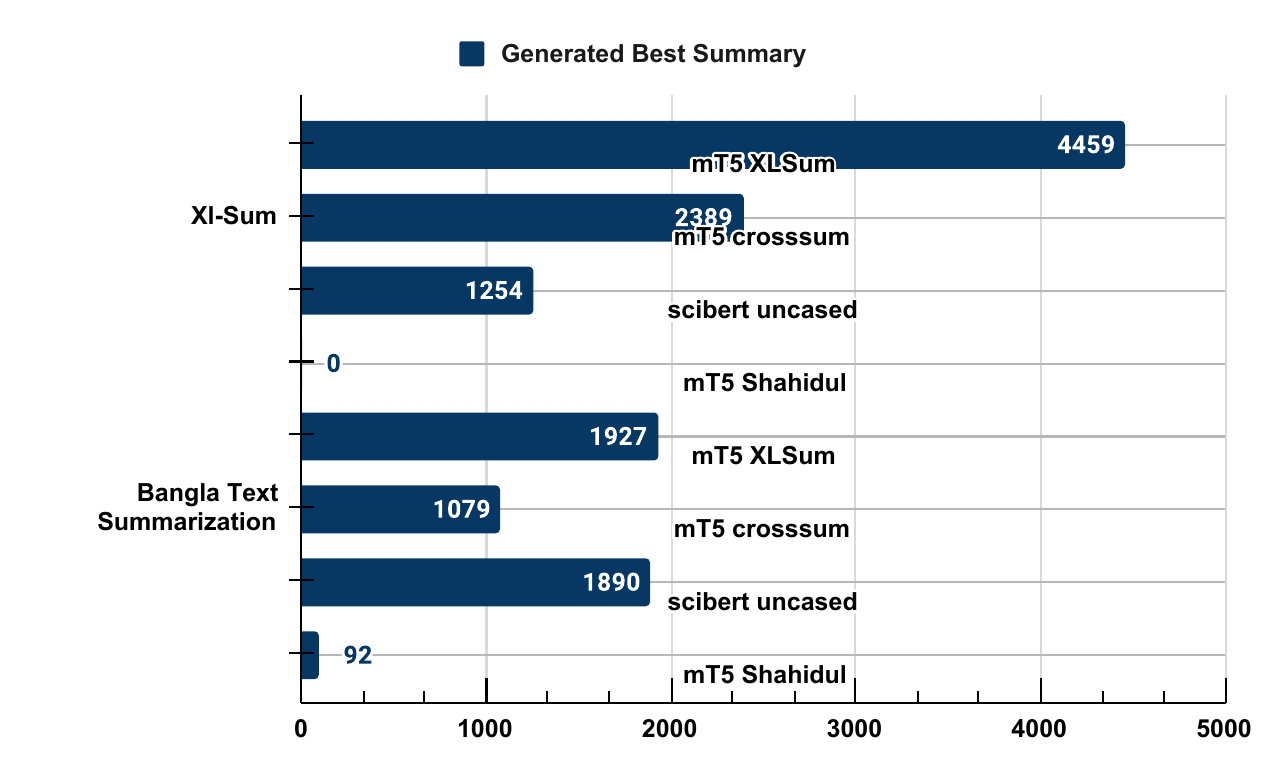}
    \caption{Statistics of the summaries per model that are selected by our approach on both datasets.}
    \label{fig:numberOfSummaries}
\end{center}
\end{figure*}
Similar scenario can be observed in the BLEU (3,4) and ROUGE scores, as shown in Table \ref{tab:rogue_bleu} for both datasets. As anticipated, our top-ranked summary outperforms the others in these metrics. Specifically, for the ROUGE-L score, which places our proposed approach as the second-best, the precision and F1 score exhibit higher values for the \textit{mT5 XLSum} summarizer. This model was specially trained on the \textit{XL-Sum} dataset, which we used in our study. It is important to note that ROUGE-L is slightly influenced by the summary of the \textit{mT5 XLSum} model for the precision and F1 score, as ROUGE uses LCS (Longest Common Subsequence) for calculation. However, for all other scores, summaries generated by our proposed approach consistently performed better. It is worth mentioning that the influence we observed earlier for the \textit{mT5 XLSum} model is not reflected in the scores on \textit{Bangla Text Summarization} dataset, as this dataset is unknown to the model. However, when considering the Word Error Rate (WER) score from Table \ref{tab:score_givenSummary}, our proposed approach ranks third. This discrepancy may arise due to differences in the length of the reference and candidate summaries, and we acknowledge this issue.

\noindent\textbf{(c) Statistical Analysis:} 
In order to determine the model that produces the best summaries for the two datasets, we conducted an analysis of the quantity of generated summaries. The statistics are presented in Figure \ref{fig:numberOfSummaries}. Upon analyzing the quantities from the figure, we observed that in most cases (4459 times for the \textit{XL-Sum} dataset and 1927 times for the \textit{Bangla Text Summarization} dataset), the \textit{mT5 XLSum} summarizer performed the best in generating summaries compared to the other three models. Interestingly, the \textit{mT5 by shahidul} model consistently failed to generate the best summaries for the \textit{XL-Sum} dataset. It only produced 92 summaries that were deemed the best for the \textit{Bangla Text Summarization} dataset, indicating its limited capability of this summarizer in generating suitable summaries.

\begin{figure*}[h!]
\begin{center}
    \centering
    \includegraphics[width=12cm]{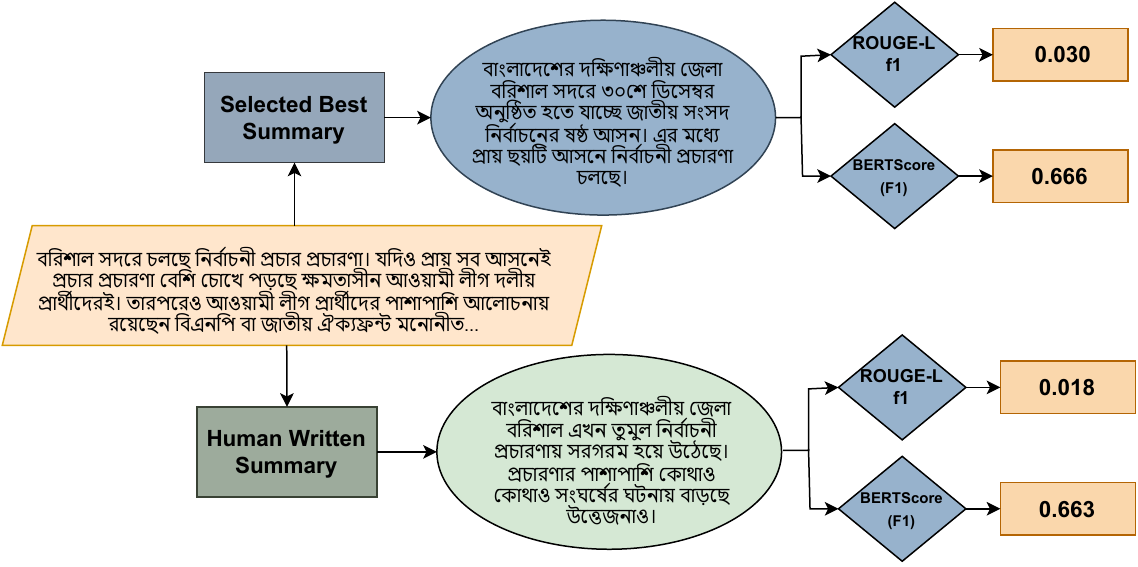}
    \caption{Performance comparison between the summary selected by our proposed approach and the reference summary on a sample text instance.}
    \label{fig:ScorePErformance}
\end{center}
\end{figure*}

\subsection{Qualitative Analysis}
In this section, we provide an example that compares a human-annotated summary with the selected best summary using our proposed approach. We present the example in Figure \ref{fig:ScorePErformance}. The input consists of a text, and the outputs include both the human-annotated summary and the summary generated by our proposed approach, along with their respective ROUGE-L (F1-measure) and BERTScore (F1-measure) scores. When comparing the original text to the summary, the proposed approach generated summary achieves a ROUGE-L (F1) score of 0.030, while the human-annotated summary achieves a score of 0.018. This slightly improved result demonstrates the capability of the proposed approach to generate a reliable and more precise summary.

\section{Related Works}
Text summarization is an important task in natural language processing, and researchers have explored various approaches to tackle this problem. Sumya et al. \cite{ref9} employed TF-IDF to calculate word and sentence scores, followed by sentence clustering to generate summaries. The use of pretrained models for text generation is a growing trend. Rahul et al. \cite{ref8} explored pre-trained models and conducted several data manipulation tasks applied to the ILSUM 2022 Indian language datasets. Ranking based text summarization has to be focused on various methodologies in text ranking. Measurement of sentence similarities comparing with a reference is the key methodology of text ranking. Various methodologies have been proposed to measure sentence similarity, such as those described by Kaisar et al. \cite{ref1} for Bengali sentence similarity measurement. They used different algorithms and computed the WMD between two Bengali summary sentences to compare their differences. Shibli et al. \cite{ref2} developed text ranking based pipeline to automatically back-transliterate Romanized Bengali into Bengali using nine free source back transliteration tools. They generated nine candidate transliterations from a single input text and selected the best one using a ranking algorithm based on a BERT-based sentence similarity graph. Palakorn et al. \cite{ref4} assessed the efficiency of different sentence similarity measures, including BERT. Additionally, Jinhyuk et al. \cite{ref5}, Jingtao et al. \cite{ref7}, and Rodrigo et al. \cite{ref8} used various BERT-based models for ranking paragraphs and documents.

\section{Conclusion}
Text summarization is a valuable tool for condensing large amounts of text and extracting key information. While low-resource languages like Bengali pose unique challenges for text summarization, the advent of pre-trained transformer models has significantly improved the accuracy and effectiveness of Bengali text summarization systems. By employing a rank-based approach that leverages multiple models and selects the best summary, the accuracy and quality of the generated summaries can be enhanced. We believe that this study will help the researchers to explore the potential of such an approach in the context of Bengali text summarization and contribute to the advancement of natural language processing techniques in low-resource language settings.

\bibliographystyle{splncs03_unsrt}
\bibliography{reference}

\begin{thebibliography}{10}
\providecommand{\url}[1]{\texttt{#1}}
\providecommand{\urlprefix}{URL }

\bibitem{nenkova2012survey}
Nenkova, A., McKeown, K.: A survey of text summarization techniques. Mining
  text data pp. 43--76 (2012)

\bibitem{kumar2021study}
Kumar, Y., Kaur, K., Kaur, S.: Study of automatic text summarization approaches
  in different languages. Artificial Intelligence Review  54(8),  5897--5929
  (2021)

\bibitem{uddin2007study}
Uddin, M.N., Khan, S.A.: A study on text summarization techniques and implement
  few of them for bangla language. In: 2007 10th international conference on
  computer and information technology. pp. 1--4. IEEE (2007)

\bibitem{ref12}
Hasan, T., Bhattacharjee, A., Islam, M.S., Mubasshir, K., Li, Y.F., Kang, Y.B.,
  Rahman, M.S., Shahriyar, R.: {XL}-sum: Large-scale multilingual abstractive
  summarization for 44 languages. In: Findings of the Association for
  Computational Linguistics: ACL-IJCNLP 2021. Association for Computational
  Linguistics (Aug 2021)

\bibitem{ref15}
Devlin, J., Chang, M.W., Lee, K., Toutanova, K.: {BERT}: Pre-training of deep
  bidirectional transformers for language understanding. In: Proceedings of the
  2019 Conference of the North {A}merican Chapter of the Association for
  Computational Linguistics: Human Language Technologies, Volume 1 (Long and
  Short Papers). Association for Computational Linguistics (Jun 2019)

\bibitem{ref11}
Xue, L., Constant, N., Roberts, A., Kale, M., Al-Rfou, R., Siddhant, A., Barua,
  A., Raffel, C.: m{T}5: A massively multilingual pre-trained text-to-text
  transformer. In: Proceedings of the 2021 Conference of the North American
  Chapter of the Association for Computational Linguistics: Human Language
  Technologies. Association for Computational Linguistics (Jun 2021)

\bibitem{pan2009survey}
Pan, S.J., Yang, Q.: A survey on transfer learning. IEEE Transactions on
  knowledge and data engineering  22(10),  1345--1359 (2009)

\bibitem{liu-lapata-2019-text}
Liu, Y., Lapata, M.: Text summarization with pretrained encoders. In:
  Proceedings of the 2019 Conference on Empirical Methods in Natural Language
  Processing and the 9th International Joint Conference on Natural Language
  Processing (EMNLP-IJCNLP). Association for Computational Linguistics (Nov
  2019)

\bibitem{b15}
Mihalcea, R., Tarau, P.: Textrank: Bringing order into text. In: Proceedings of
  the 2004 conference on empirical methods in natural language processing. pp.
  404--411 (2004)

\bibitem{b16}
Page, L., Brin, S., Motwani, R., Winograd, T.: The pagerank citation ranking:
  Bringing order to the web. Tech. rep., Stanford InfoLab (1999)

\bibitem{ref2}
Shibli, G.S., Shawon, M.T.R., Nibir, A.H., Miandad, M.Z., Mandal, N.C.:
  Automatic back transliteration of romanized bengali (banglish) to bengali.
  Iran Journal of Computer Science pp. 1--12 (2022)

\bibitem{ref16}
Bhattacharjee, A., Hasan, T., Ahmad, W., Mubasshir, K.S., Islam, M.S., Iqbal,
  A., Rahman, M.S., Shahriyar, R.: {B}angla{BERT}: Language model pretraining
  and benchmarks for low-resource language understanding evaluation in
  {B}angla. In: Findings of the Association for Computational Linguistics:
  NAACL 2022. Association for Computational Linguistics (Jul 2022)

\bibitem{b18}
Han, J., Kamber, M., Pei, J., et~al.: Getting to know your data. In: Data
  mining. vol.~2, pp. 39--82. Morgan Kaufmann Boston, MA, USA (2012)

\bibitem{ref13}
Bhattacharjee, A., Hasan, T., Ahmad, W.U., Li, Y.F., Kang, Y.B., Shahriyar, R.:
  {C}ross{S}um: Beyond {E}nglish-centric cross-lingual summarization for 1,500+
  language pairs. In: Proceedings of the 61st Annual Meeting of the Association
  for Computational Linguistics (Volume 1: Long Papers). Association for
  Computational Linguistics (Jul 2023)

\bibitem{ref14}
Beltagy, I., Lo, K., Cohan, A.: {S}ci{BERT}: A pretrained language model for
  scientific text. In: Proceedings of the 2019 Conference on Empirical Methods
  in Natural Language Processing and the 9th International Joint Conference on
  Natural Language Processing (EMNLP-IJCNLP). Association for Computational
  Linguistics (Nov 2019)

\bibitem{b19}
Papineni, K., Roukos, S., Ward, T., Zhu, W.J.: Bleu: a method for automatic
  evaluation of machine translation. In: Proceedings of the 40th annual meeting
  of the Association for Computational Linguistics. pp. 311--318 (2002)

\bibitem{b20}
Lin, C.Y.: Rouge: A package for automatic evaluation of summaries. In: Text
  summarization branches out. pp. 74--81 (2004)

\bibitem{b21}
Zhang*, T., Kishore*, V., Wu*, F., Weinberger, K.Q., Artzi, Y.: Bertscore:
  Evaluating text generation with bert. In: International Conference on
  Learning Representations (2020),
  \url{https://openreview.net/forum?id=SkeHuCVFDr}

\bibitem{b23}
Banerjee, S., Lavie, A.: Meteor: An automatic metric for mt evaluation with
  improved correlation with human judgments. In: Proceedings of the acl
  workshop on intrinsic and extrinsic evaluation measures for machine
  translation and/or summarization. pp. 65--72 (2005)

\bibitem{b24}
Morris, A.C., Maier, V., Green, P.: From wer and ril to mer and wil: improved
  evaluation measures for connected speech recognition. In: Eighth
  International Conference on Spoken Language Processing (2004)

\bibitem{ref9}
Akter, S., Asa, A.S., Uddin, M.P., Hossain, M.D., Roy, S.K., Afjal, M.I.: An
  extractive text summarization technique for bengali document (s) using
  k-means clustering algorithm. In: 2017 IEEE International Conference on
  Imaging, Vision \& Pattern Recognition (icIVPR). pp. 1--6. IEEE (2017)

\bibitem{ref8}
Tangsali, R., Pingle, A., Vyawahare, A., Joshi, I., Joshi, R.: Implementing
  deep learning-based approaches for article summarization in indian languages.
  arXiv preprint arXiv:2212.05702  (2022)

\bibitem{ref1}
Masum, A.K.M., Abujar, S., Tusher, R.T.H., Faisal, F., Hossain, S.A.: Sentence
  similarity measurement for bengali abstractive text summarization. In: 2019
  10th International Conference on Computing, Communication and Networking
  Technologies (ICCCNT). pp. 1--5. IEEE (2019)

\bibitem{ref4}
Achananuparp, P., Hu, X., Shen, X.: The evaluation of sentence similarity
  measures. In: Data Warehousing and Knowledge Discovery: 10th International
  Conference, DaWaK 2008 Turin, Italy, September 2-5, 2008 Proceedings 10. pp.
  305--316. Springer (2008)

\bibitem{ref5}
Lee, J., Yun, S., Kim, H., Ko, M., Kang, J.: Ranking paragraphs for improving
  answer recall in open-domain question answering. In: Proceedings of the 2018
  Conference on Empirical Methods in Natural Language Processing. Association
  for Computational Linguistics (Oct-Nov 2018)

\bibitem{ref7}
Nogueira, R., Yang, W., Cho, K., Lin, J.: Multi-stage document ranking with
  bert. arXiv preprint arXiv:1910.14424  (2019)

\end{thebibliography}



\end{document}